\documentclass[11pt,a4paper]{article}
\usepackage[hyperref]{emnlp-ijcnlp-2019}
\usepackage{times}
\usepackage{latexsym}

\usepackage{url}

\usepackage{amsfonts}
\usepackage{bm}
\usepackage{array,amsmath}
\usepackage{amssymb}
\usepackage{dsfont}
\usepackage{float}
\usepackage{graphicx}
\usepackage{algorithm}
\usepackage{algorithmicx}
\usepackage{algpseudocode}
\usepackage{pgf,tikz}
\usepackage{mathrsfs}
\usepackage{multirow}
\usetikzlibrary{arrows}

\def\figref#1{Figure~\ref{fig:#1}}
\def\figlabel#1{\label{fig:#1}\label{p:#1}}

\def\tabref#1{Table~\ref{tab:#1}}
\def\tablabel#1{\label{tab:#1}\label{p:#1}}

\def\secref#1{\S\ref{sec:#1}}
\def\seclabel#1{\label{sec:#1}}
\def\eqref#1{Eq.~\ref{eqn:#1}}

\def\eqlabel#1{\label{eqn:#1}}

\def\method{DensRay}
\def\analogymethod{IntCos}

\aclfinalcopy %

\title{Analytical Methods for Interpretable Ultradense Word Embeddings}

\author{Philipp Dufter, Hinrich Sch\"{u}tze\\ Center for Information and
	Language Processing (CIS) \\ LMU Munich, Germany\\ {\tt philipp@cis.lmu.de}}

\date{}

\begin{document}
\maketitle 

\begin{abstract} 
	
	Word embeddings are useful for a
	wide variety of tasks, but they lack interpretability.  By rotating word
	spaces, interpretable dimensions can be identified while preserving the
	information contained in the embeddings without any loss. In this work, we
	investigate three methods for making word spaces interpretable by rotation:
	Densifier \cite{rothe2016ultradense}, linear SVMs and \method{}, a new method
	we propose.  In contrast to Densifier, \method{} can
	be computed in closed form, is hyperparameter-free and thus more robust than
	Densifier.  We evaluate the three methods on lexicon induction and set-based word
	analogy. In addition we provide qualitative insights
	as to how interpretable word spaces can be used for removing gender bias from embeddings.
	
\end{abstract}

\section{Introduction}

Distributed representations for words have been of interest in natural language
processing for many years. Word embeddings have been particularly effective and
successful. 
On the downside, embeddings are generally not
interpretable. But interpretability
is desirable
for several reasons. i) Semantically or syntactically similar words can be
extracted: e.g., for lexicon induction.
ii) Interpretable dimensions can be used to evaluate word spaces by examining which information is covered by the embeddings. iii)
Computational advantage: for a high-quality sentiment classifier only a couple
of dimensions of a high-dimensional word space are
relevant. iv) By removing interpretable dimensions one can remove unwanted
information (e.g., gender bias).  v) Most importantly, interpretable embeddings
support the goal of interpretable deep learning models.

Orthogonal transformations have been of particular interest in the
literature. The reason is twofold: under the assumption that existing word
embeddings are of high-quality one would like to preserve the original
embedding structure by using orthogonal transformations (i.e., preserving
original distances). \newcite{park2017rotated} provide evidence that rotating
existing dense word embeddings achieves the best performance across a range of
interpretability tasks. 

In this work we modify  the objective function of
Densifier \cite{rothe2016ultradense} such that a closed form solution becomes
available. We call this method \method{}. Following
\newcite{amir2015inesc} we compute simple linear SVMs, which we find to perform
surprisingly well. We compare these methods on the task of lexicon induction.

Further, we show how interpretable word spaces can be
applied to other tasks: first we 
use interpretable word spaces for debiasing
embeddings. Second we show how they can be used for solving
the set-based word analogy task. To this end, we introduce
the set-based method  \analogymethod{}, which is
closely related to LRCos introduced by
\newcite{drozd2016word}. We find \analogymethod{} to perform
comparable to LRCos, but to be preferable for analogies
which are hard to solve.

Our contributions are: \textbf{i)} We
modify Densifier's objective function and derive an analytical solution for computing interpretable embeddings.
\textbf{ii)} We show that the analytical solution performs as well as Densifier but
is more robust. \textbf{iii)} We provide evidence that simple linear SVMs are best
suited for the task of lexicon induction.  \textbf{iv)} We demonstrate how interpretable
embedding spaces can be used for debiasing embeddings and solving the set-based word analogy task. The source code of our experiments is available.\footnote{\url{https://github.com/pdufter/densray}}

\section{Methods}

\subsection{Notation}
We consider a vocabulary $V := \{v_1, v_2, ..., v_n\}$
together with an embedding matrix $E \in \mathbb{R}^{n
  \times d}$ where $d$ is the embedding dimension. The $i$th
row of $E$ is the vector $e_i$.\footnote{We denote the
  vector corresponding to a word $w$ by $e_w$.} We require an
annotation for a specific linguistic feature (e.g.,
sentiment) and denote this annotation by $l:V \to \{-1,
1\}$. The objective is to find an orthogonal matrix $Q \in
\mathbb{R}^{d \times d}$ such that $EQ$ is interpretable,
i.e., the values of the first $k$ dimensions correlate well
with the linguistic feature. We refer to the first $k$
dimensions as interpretable ultradense word space. We
interpret $x \in \mathbb{R}^n$ as a column vector and
$x^\intercal$ as a row vector. Further, we normalize all word embeddings with respect to the euclidean norm.

\subsection{\method{}}
Throughout this section $k=1$. Given a linguistic signal $l$
(e.g., sentiment), consider  $L_=:= \{(v, w) \in V \times V |\; l(v) = l(w)\}$, and analogously $L_{\not =}$. We call $d_{vw}:= e_v - e_w$ a difference vector.

Densifier \cite{rothe2016ultradense} solves the following optimization problem,
\begin{align*}
\max_q \sum _ {(v,w) \in L_{\not =}} \alpha_{\not =}  &\left\| q^\intercal d_{vw} \right\| _ { 2 } - \\
&\sum _ { (v,w) \in L_{=}} \alpha_{=}  \left\| q^\intercal d_{vw} \right\|_ { 2 }, 
\end{align*}
subject to $q^\intercal q = 1$ and $q \in \mathbb{R}^d$. Further $\alpha_{\not =}, \alpha_{=}  \in [0, 1]$ are hyperparameters. We now modify
the objective function: we use the squared euclidean norm instead of the euclidean norm, something that is frequently done in optimization to simplify the gradient. The problem becomes then 
\begin{align}
\max_q \sum _ {(v,w) \in L_{\not =}} \alpha_{\not =} &\left\| q^\intercal d_{vw} \right\| _ { 2 }^2- \nonumber \\
&\sum _ { (v,w) \in L_{=}}  \alpha_{=}  \left\| q^\intercal d_{vw} \right\| _ { 2 }^2. \eqlabel{cosine_opt}
\end{align}
Using $\left\|x\right\|_2^2 = x^\intercal x$ together with associativity of the matrix product we can simplify to
\begin{align}\eqlabel{mainopt}
\max_{q}\; & q^\intercal \Big( \alpha_{\not =}  \sum_{(v,w) \in L_{\not =}} d_{vw}d_{vw}^\intercal - \\
&\qquad\qquad \alpha_{=}  \sum_{(v,w) \in L_{=}} d_{vw}d_{vw}^\intercal \Big) q \nonumber \\ 
& =:\; \max_{q}\; q^\intercal  A q  \quad \text{subject to } q^\intercal q = 1. \nonumber
\end{align}

Thus we aim to maximize the Rayleigh quotient of $A$ and $q$. Note that $A$ is a real symmetric matrix. Then it is well known that the eigenvector belonging to the maximal eigenvalue of $A$ solves the above problem (cf. \newcite[Section 4.2]{horn1990matrix}). We call this analytical solution \textbf{\method{}}.

A second dimension that is orthogonal to the first dimension and encodes the
linguistic features second strongest is given by the eigenvector corresponding to the 
second largest eigenvalue. The matrix of $k$ eigenvectors of $A$ ordered by  the corresponding eigenvalues yields  the desired matrix $Q$ (cf. \newcite[Section 4.2]{horn1990matrix}) for $k>1$. Due to $A$ being a real symmetric matrix, $Q$ is always orthogonal.

\subsection{Comparison to Densifier}

We have shown that \method{} is a closed form solution
to our new formalization of Densifier. This formalization entails differences. 

\textbf{Case $k>1$.}
While both methods -- Densifier and \method{} -- yield ultradense $k$ dimensional subspaces. While we show that the spaces are comparable for $k = 1$ we leave it to future work to examine how the subspaces differ for $k > 1$. 

\textbf{Multiple linguistic signals.}
Given multiple linguistic features, Densifier can obtain a single orthogonal transformation simultaneously for all linguistic features with chosen dimensions reserved for different features. \method{} can encode multiple linguistic features in one transformation only by iterative application.

\textbf{Optimization.}
Densifier is based on solving an optimization problem using stochastic gradient descent  with iterative orthogonalization of $Q$. \method{}, in contrast, is an analytical solution. Thus we expect \method{} to be more robust, which is confirmed by our experiments. 

\subsection{Geometric Interpretation}

Assuming we normalize the vectors $d_{vw}$ one can interpret
\eqref{cosine_opt} as follows: we search for a unit vector
$q$ such that the square of the cosine similarity with
$d_{vw}$ is large if $(v, w) \in L_{\not =}$ and small if $(v, w) \in L_{=}$. Thus, we identify dimensions that are parallel/orthogonal to
difference vectors of words belonging to different/same classes. It seems reasonable to consider the average cosine similarity. Thus if $n_=$, $n_{\not =}$ is the number of elements in $L_{=}$, $L_{\not =}$ one can choose $\alpha_{\not =}  = n_{\not =}^{-1}$ and $\alpha_{=} = n_{=}^{-1}$. 

\section{Lexicon Induction}

We show that \method{} and Densifier indeed perform
comparably using the task of lexicon induction. We adopt
\newcite{rothe2016ultradense}'s experimental setup.  We also
use \newcite{rothe2016ultradense}'s code for Densifier.  Given
a word embedding space and a sentiment/concreteness
dictionary (binary or continuous scores where we binarize
continuous
  scores using the median), we identify a
one-dimensional interpretable subspace. Subsequently we use
the values along this dimension to predict a score for
unseen words and report Kendall's $\tau$ rank correlation with
the gold scores.

To ensure comparability across methods we have redone all
experiments in the same setting: we
deduplicated lexicons, removed a potential train/test
overlap and ignored neutral words in the lexicons. We set $\alpha_{\not =} =  \alpha_{=} =0.5$ to ensure comparability between Densifier and \method{}.

Additionally we report results created by linear SVM/SVR inspired be their good performance as demonstrated by
\newcite{amir2015inesc}. While they did not use linear kernels, we require linear kernels to obtain interpretable dimensions. Naturally the normal vector of the hyperplane in SVMs/SVRs reflects an interpretable dimension. An orthogonal transformation can be computed by considering a random orthogonal basis of the null space of the interpretable dimension.

\tabref{lexica} shows results.  As expected the performance of
Densifier  and \method{} is comparable (macro mean deviation
of 0.001). We explain slight deviations between the results
with the slightly different objective functions of \method{}
and Densifier. In addition, the re-orthogonalization used in Densifier can result in an unstable training process.  \figref{mean}
assesses  the stability
by reporting mean and standard deviation
for the concreteness task (BWK lexicon). We varied
the size of the training lexicon as depicted on the x-axis
and sampled 40 subsets of the lexicon with the prescribed
size. For the sizes 512 and 2048 Densifier shows an increased standard deviation. 
This is because there is at least one sample for which the performance significantly drops. 
Removing the re-orthogonalization in Densifier prevents the drop and restores performance. Recent work \cite{zhao2019multilingual} also finds that replacing the orthogonalization with a regularization is reasonable in certain circumstances. Given that \method{} and Densifier yield the same performance and \method{} is a stable closed form solution always yielding a orthogonal transformation we conclude that \method{} is preferable.

Surprisingly, simple linear SVMs perform best in the task of lexicon induction. SVR is slightly better when continuous lexica are used for training (line 8). Note that the eigendecomposition used in DensRay 
yields a basis with dimensions ordered by their correlation with the linguistic feature. An SVM can achieve this only by iterated application.

\begin{table}[H]
	\centering
	\tiny
	\def\sep{0.075cm}
	\begin{tabular}{l@{\hspace{\sep}}l@{\hspace{\sep}}l@{\hspace{\sep}}l@{\hspace{\sep}}l@{\hspace{\sep}}|rrrr}
		&Task&Emb.& Lex. (Train) & Lex. (Test) & Dens. & \method{} & SVR &SVM\\
		\hline
		1 & sent & CZ  & SubLex & SubLex & 0.546& 0.549& \textbf{0.585}& 0.585 \\
		2 & sent & DE  & GermanPC & GermanPC & 0.636& 0.631& 0.674& \textbf{0.677} \\
		3 & sent & ES  & fullstrength & fullstrength & 0.541& 0.546& 0.571& \textbf{0.576} \\
		4 & sent & FR  & FEEL & FEEL & 0.469& 0.471& 0.555& \textbf{0.565} \\
		5 & sent & EN  & WHM & WHM & 0.623& 0.623& \textbf{0.627}& 0.625 \\
		6 & sent & EN(t)  & WHM & SE Trial* & 0.624& 0.621& 0.618& \textbf{0.637} \\
		7 & sent & EN(t) &  WHM & SE Test*  & 0.600& 0.608& 0.619& \textbf{0.636} \\
		8 & conc & EN  & BWK* & BWK* & 0.599& 0.602& \textbf{0.655}& 0.641 \\
		\hline
		9 &  \multicolumn{4}{c}{Macro Mean}& 0.580& 0.581& 0.613& \textbf{0.618}
	\end{tabular}
	\caption{Results on lexicon induction. Numbers are Kendall $\tau$ rank correlation. For details on the resources see \tabref{lexindres} and \cite{rothe2016ultradense}. Bold: best result across methods. $^\star$: continuous lexicon.}
	\tablabel{lexica}
\end{table}

\begin{table}[h]
	\tiny
	\centering
	\begin{tabular}{lp{5.5cm}}
		Name & Description \\
		\hline
		CZ, DE, ES & Czech, German, Spanish embeddings by \cite{rothe2016ultradense} \\
		FR & French frWac embeddings \cite{fauconnier_2015} \\
		EN & English GoogleNews embeddings \cite{mikolov2013efficient} \\
		EN(t) & English Twitter Embeddings \cite{rothe2016ultradense} \\
		& \\
	\end{tabular}
	\begin{tabular}{lp{5.5cm}}
		Name & Description \\
		\hline
		SubLex & Czech sentiment lexicon \cite{veselovska2013czech} \\
		GermanPC & German sentiment lexicon  \cite{waltinger2010germanpolarityclues}\\
		fullstrength & Spanish sentiment lexicon \cite{perez2012learning}\\
		FEEL & French sentiment lexicon \cite{abdaoui2017feel} \\
		WHM & English sentiment lexicon; combination of MPQA \cite{wilson2005recognizing}, Opinion Lexicon \cite{hu2004mining} and NRC emotion lexcion \cite{mohammad2013crowdsourcing}\\
		SE & Semeval 2015 Task 10E shared task data \cite{rosenthal2015semeval}\\
		BWK & English concreteness lexicon \cite{brysbaert2014concreteness} \\
	\end{tabular}
	\caption{Overview of resources for lexicon induction. The setup is identical to \cite{rothe2016ultradense}.}\tablabel{lexindres}
\end{table}

\begin{figure}
	\centering
	\includegraphics[width=0.75\linewidth]{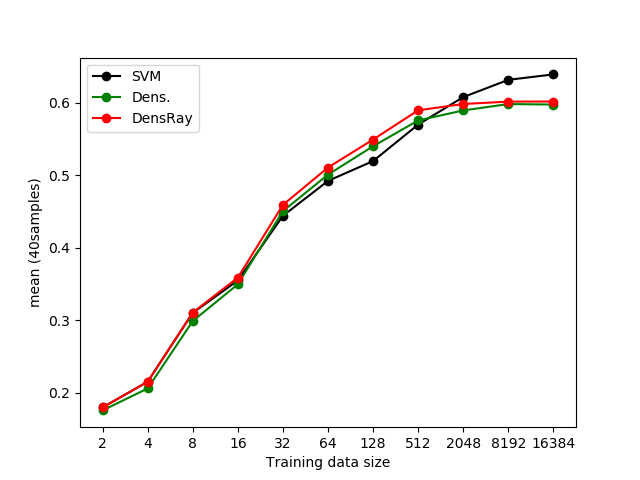}
	\includegraphics[width=0.75\linewidth]{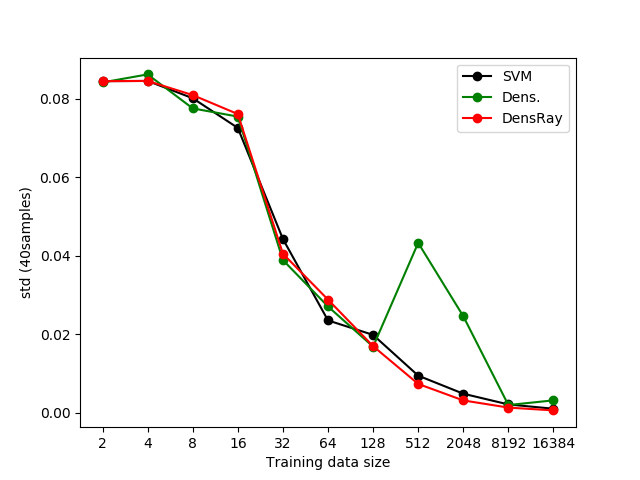}
	\caption{Mean (top) and standard deviation (bottom) of the performance across 40 samples of the training lexicon with varying sizes. Performed on the English concreteness task (line 8 in \tabref{lexica}). SVR performs similar to SVM and is omitted for clarity.}
	\figlabel{mean}
\end{figure}

\section{Removing Gender Bias}\seclabel{gender}

Word embeddings are well-known for encoding prevalent biases
and stereotypes (cf.\ \newcite{bolukbasi2016man}). We
demonstrate qualitatively that by identifying an interpretable gender
dimension and subsequently removing this dimension, one can
remove parts of gender information that potentially could
cause biases in downstream processing. Given the original word space $E$ we consider the interpretable space $E':= EQ$, where $Q$ is computed using \method{}. We denote by $E_{\cdot, -1}  \in \mathbb{R}^{n \times(d-1)}$ the word space with removed first dimension and call it the ``complement'' space. We expect $E_{\cdot, -1}$ to be a word space with less gender bias.

To examine this approach qualitatively we use a list of occupation names\footnote{\tiny{\url{https://github.com/tolga-b/debiaswe/blob/master/data/professions.json}} \nopagebreak } by \newcite{bolukbasi2016man} and examine the cosine similarities of occupations with the vectors of ``man'' and ``woman''. \figref{debiasscatter} shows the similarities in the original space $E$ and debiased space $E_{\cdot, -1} $. One can see the similarities are closer to the identity (i.e., same distance to ``man'' and ``woman'') in the complement space. To identify occupations with the greatest bias, \tabref{debias} lists occupations for which $\text{sim}(e_w, e_{\text{man}}) - \text{sim}(e_w, e_{\text{woman}})$ is largest/smallest. One can clearly see a debiasing effect when considering the complement space. Extending this qualitative study to a more rigorous quantitative evaluation is part of future work. 

\begin{figure}
	\centering
	\includegraphics[width=.65\linewidth]{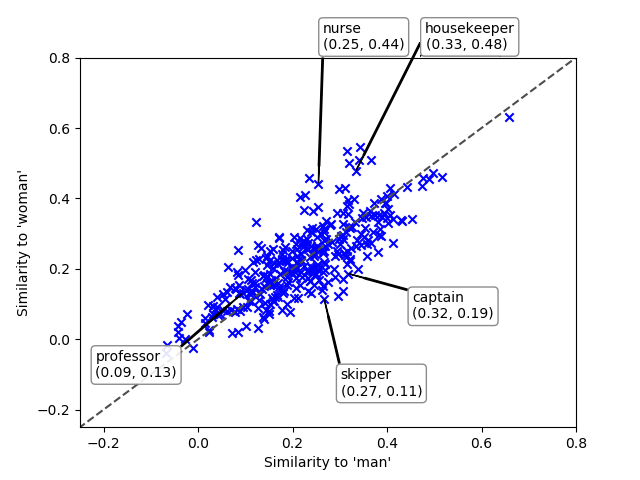}
	\includegraphics[width=.65\linewidth]{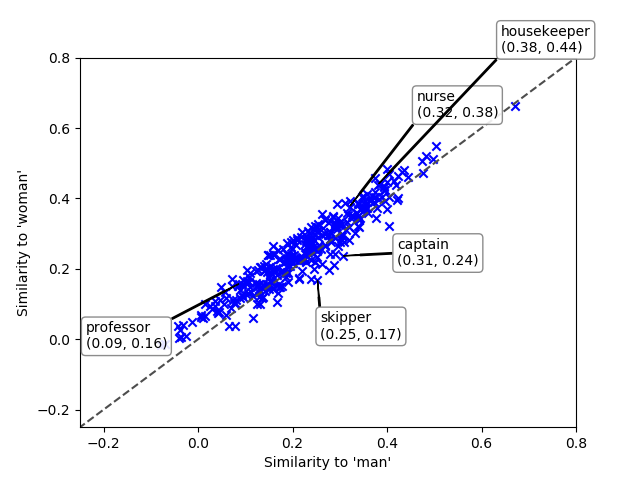}
	\caption{Similarities of occupation vectors with the vectors of man and woman. Top shows the original word space and bottom the word space with removed gender dimension.}
	\figlabel{debiasscatter}
\end{figure}

\begin{table}[H]
	\centering
	\tiny
	\begin{tabular}{l|lrr||lrr}
		& \multicolumn{3}{c}{Original Space} & \multicolumn{3}{c}{Complement Space}  \\
		& & man & woman && man & woman \\
		\hline
		
		\multirow{5}{0.2cm}{\rotatebox{90}{female bias}} &
		actress & 0.23 & 0.46 &  lawyer & 0.16 & 0.27 \\ 
		&businesswoman & 0.32 & 0.53 &  ambassador & 0.07 & 0.17 \\ 
		&registered\_nurse & 0.12 & 0.33 &  attorney & 0.05 & 0.15 \\ 
		&housewife & 0.34 & 0.55 &  legislator & 0.26 & 0.36 \\ 
		&homemaker & 0.22 & 0.40 &  minister & 0.10 & 0.20 \\ 
		\hline
		& \multicolumn{6}{c}{...}\\
		\hline
		\multirow{5}{0.2cm}{\rotatebox{90}{male bias}} &   hitman & 0.41 & 0.27 &  captain &0.31 & 0.24 \\ 
		&gangster & 0.34 & 0.20 &  marksman &0.29 & 0.21 \\ 
		&skipper & 0.27 & 0.11 &  maestro &0.28 & 0.20 \\ 
		&marksman & 0.31 & 0.14 &  hitman &0.40 & 0.32 \\ 
		&maestro & 0.30 & 0.12 &  skipper &0.25 & 0.17 \\ 
	\end{tabular}
	\caption{Top 5 occupations that exhibit the greatest bias (measured by difference in cosine similarity). Numbers indicate cosine similarity between word vectors.}
	\tablabel{debias}
\end{table}

\section{Word Analogy}

In this section we use interpretable word spaces for set-based word analogy. Given a list of analogy pairs [($a$, $a'$), ($b$, $b'$),  ($c$, $c'$), \dots] the task is to predict $a'$ given $a$. \newcite{drozd2016word} provide a detailed overview over different methods, and find that their method LRCos performs best.

LRCos assumes two classes: all left elements of a pair (``left class'') and all right elements (``right class''). They train a logistic regression (LR) to differentiate between these two classes. The predicted score of the LR multiplied by the cosine similarity in the word space is their final score. Their prediction for $a'$ is the word with the highest final score.

We train the classifier on all analogy pairs except for a single pair for which we then obtain the predicted score. In addition we ensure that no word belonging to the test analogy is used during training (splitting the data only on word analogy pairs is not sufficient).

Inspired by LRCos we use interpretable word spaces for
approaching word analogy:  we train \method{} or an SVM to
obtain interpretable embeddings $E' = EQ$ using the class
information as reasoned above. We use a slightly different
notation in this section: for a word $w$ the $i$th component
of its embedding is given by $E_{w, i}$. Therefore we denote
as $E_{\cdot, 1}$ the first column of $E'$ (i.e., the most
interpretable dimension). We min-max normalize  $E_{\cdot,
  1}$
such that words belonging to the right class have a high value (i.e., we flip the sign if necessary). 
For a query word $a$ we now want to identify the corresponding $a'$ by solving
$$
\hat{a} = \arg \max_{v \in V} \; \text{norm}(E_{v, 1}) \, \text{sim}(E_{a, \cdot}, E_{v, \cdot})
$$
where \emph{sim} computes the cosine similarity. 

Given the result from \secref{gender} we extend the above
method by computing the cosine similarity in the orthogonal
complement, i.e., $\text{sim}(E_{a, -1}, E_{v, -1})$. We
call this method \textbf{\analogymethod{}} (INTerpretable, \mbox{COSine}). Depending on the space used for computing the cosine similarity add the word ``Original'' or ``Complement''.

We evaluate this method across two analogy datasets. These are the Google Analogy Dataset (GA) \cite{mikolov2013efficient} and BATS \cite{drozd2016word}. As embeddings spaces we use  Google News Embeddings (GN) \cite{mikolov2013efficient} and FastText subword embeddings (FT) \cite{bojanowski2017enriching}. We consider the first 80k word embeddings from each space.

\tabref{analogies} shows the results. The first observation is that there is no clear winner. IntCos Original performs comparably to LRCos with slight improvements for GN/BATS: here the classes are widespread and exhibit
low cosine similarity (IntraR and IntraL), which makes them harder to solve. IntCos Complement maintains performance for GN/BATS and is beneficial for Derivational analogies on GN. For most other analogies it harms performance. 

Within IntCos Original it is favorable to use DensRay as it gives slight performance improvements. Especially for harder analogies, where interclass similarity is high and intraclass similarities are low (e.g., in GN/BATS), DensRay outperforms SVMs. In contrast to SVMs,
\method{} considers difference vectors \emph{within} classes as well -- this seems to be of advantage here. 

\begin{table}
	\centering
	\tiny
	\def\sep{0.1cm}
	\def\biggersep{0.25cm}
	\begin{tabular}{c@{\hspace{\biggersep}}l@{\hspace{\sep}}||@{\hspace{\sep}}r@{\hspace{\sep}}r@{\hspace{\sep}}r@{\hspace{\sep}}||@{\hspace{\sep}}r@{\hspace{\sep}}r@{\hspace{\sep}}|@{\hspace{\sep}}r@{\hspace{\sep}}r@{\hspace{\sep}}|@{\hspace{\sep}}r@{\hspace{\sep}}}
		&& \multicolumn{3}{c}{Mean Cosine Sim} & \multicolumn{5}{c}{Precision} \\
		&& & &  & \multicolumn{4}{c}{\analogymethod{}} & LRCos \\
		&& & &  & \multicolumn{2}{c}{complement} &  \multicolumn{2}{c}{original} & \\
		&& Inter & IntraL & IntraR & DensR. & SVM & DensR. & SVM & \\
		\hline
		\hline
		\multirow{6}{0cm}{\rotatebox{90}{FT/BATS}}
		& Inflectional & 0.75& 0.48& 0.51& 0.92& 0.93&  \textbf{0.97}& \textbf{0.97}& \textbf{0.97}\\
		&Derivational & 0.63& 0.47& 0.45& 0.74& 0.78&  \textbf{0.81}& 0.80& 0.80\\
		&Encyclopedia & 0.48& 0.43& 0.55& 0.30& 0.43&  0.41& 0.43&  \textbf{0.45}\\
		&Lexicography & 0.62& 0.37& 0.38& 0.17& 0.20&  0.21& 0.22&  \textbf{0.26}\\
		&Macro Mean & 0.62& 0.44& 0.47& 0.53& 0.58& 0.60& 0.60& \textbf{0.61}\\
		&Macro Std & 0.12& 0.06& 0.09& 0.34 & 0.33& 0.34& 0.33&  \textbf{0.32}\\
		\hline
		\multirow{6}{0cm}{\rotatebox{90}{GN/BATS}}
		&Inflectional & 0.63& 0.22& 0.23& \textbf{0.88}& 0.87& \textbf{0.88}& \textbf{0.88}& \textbf{0.88}\\
		&Derivational & 0.44& 0.21& 0.20& \textbf{0.55}& 0.50& 0.51 & 0.48& 0.44\\
		&Encyclopedia & 0.35& 0.29& 0.42& 0.33& \textbf{0.35}&  \textbf{0.35}& 0.32& 0.34\\
		&Lexicography & 0.45& 0.17& 0.18& \textbf{0.19}& 0.17& \textbf{0.19}& 0.17&  0.18\\
		&Macro Mean & 0.46& 0.22& 0.26& \textbf{0.48}& 0.47& \textbf{0.48}& 0.46&  0.45\\
		&Macro Std  & 0.14& 0.07& 0.12& \textbf{0.31}& \textbf{0.31}& 0.32& 0.32&  0.32\\
		\hline
		\multirow{3}{0cm}{\rotatebox{90}{FT/GA}}
		&Micro Mean & 0.73& 0.48& 0.53& 0.88& 0.91& \textbf{0.93}& 0.92&  \textbf{0.93}\\
		&Macro Mean & 0.71& 0.50& 0.53& 0.87& 0.90&  \textbf{0.91}& 0.90& 0.89\\
		&Macro Std & 0.11& 0.05& 0.06& 0.11& \textbf{0.08}&  0.12& 0.17&  0.23\\
		\hline
		\multirow{3}{0cm}{\rotatebox{90}{GN/GA}}
		&Micro Mean& 0.62& 0.31& 0.36& 0.85& 0.87&  \textbf{0.89}& 0.87&  0.88\\
		&Macro Mean & 0.61& 0.30& 0.35& 0.85& 0.86&  \textbf{0.88}& 0.85& 0.87\\
		&Macro Std & 0.10& 0.09& 0.10& 0.08& \textbf{0.07}& 0.09& 0.11&  0.11\\
		\hline
	\end{tabular}
	\caption{Left part shows mean cosine similarity. Inter: mean cosine similarity between pairs. IntraL/R: mean cosine similarity within the left/right class. Right part shows precision for word analogy task.}
	
	\tablabel{analogies}
\end{table}

\section{Related Work}

\textbf{Identifying Interpretable Dimensions.} Most relevant to our method
is a line of work that uses transformations of existing word spaces to obtain
interpretable subspaces. \newcite{rothe2016ultradense} compute an orthogonal
transformation using shallow neural networks. \newcite{park2017rotated} apply
exploratory factor analysis to embedding spaces to obtain interpretable
dimensions in an unsupervised manner. Their approach relies on solving complex
optimization problems, while we focus on closed form solutions. \newcite{senel2018semantic} use SEMCAT categories in
combination with the Bhattacharya distance to identify interpretable
directions. Also, oriented PCA \cite{diamantaras1996principal} is closely
related to our method. However, both methods yield non-orthogonal
transformation. \newcite{faruqui2015retrofitting} use semantic lexicons to
retrofit embedding spaces. Thus they do not fully maintain the structure of the
word space, which is in contrast to this work.

\textbf{Interpretable Embedding Algorithms.}  Another line of work modifies
embedding algorithms to yield interpretable dimensions \cite{kocc2018imparting,luo2015online,shin2018interpreting,zhao2018learning}. There is also much work that generates sparse embeddings that are claimed to be
more interpretable \cite{murphy2012learning,faruqui2015sparse,fyshe2015compositional,subramanian2018spine}. Instead of learning new embeddings, we
aim at making dense embeddings interpretable.

\section{Conclusion}

We investigated analytical methods for obtaining interpretable word spaces. Relevant methods were examined with the tasks of lexicon induction, word analogy and debiasing.
We gratefully
\textbf{acknowledge}
funding through a
Zentrum Digitalisierung.Bayern 
fellowship awarded to
the first author. This work was supported
by the European Research Council (\# 740516). 
We thank the anonymous reviewers for valuable comments.

\bibliography{dyadic}
\bibliographystyle{acl_natbib}

\newpage 
\null
\newpage
\section{Appendix}

\subsection{Code}

The code which was used to conduct the experiments in this paper is available at \url{https://github.com/pdufter/densray}.

\subsection{Continuous Lexicon}

In case of a continuous lexicon  $l:V \to \mathbb{R}$ one can extend Equation 2 in the main paper by defining: 
\begin{equation*}
A := \sum_{(v,w) \in V \times V} - l(v)l(w) d_{vw}d_{vw}^\intercal 
\end{equation*}
In the case of a binary lexicon Equation 2 from the main paper is recovered for $\alpha_{\not =} = \alpha_{=} = 1$.

\subsection{Full Analogy Results}

In this section we present the results of the word analogy task per category.  See \tabref{complement} and \tabref{original} for detailed results with the methods \analogymethod{} Complement and Original, respectively. The format and numbers presented are the same as in the corresponding table from the main paper.

\begin{table*}
	\def\sep{0.01cm}
	\tiny
	\centering
	\begin{tabular}{ccc}
		& \textbf{FastText} &\textbf{ Google News}\\[10pt]
		
		\rotatebox{90}{\textbf{Google Analogy}} &	\begin{tabular}{l@{\hspace{\sep}}||r@{\hspace{\sep}}r@{\hspace{\sep}}r@{\hspace{\sep}}||r@{\hspace{\sep}}r@{\hspace{\sep}}|r@{\hspace{\sep}}}
			& \multicolumn{3}{c}{Mean Cosine Sim} & \multicolumn{3}{c}{Precision} \\
			& & &  & \multicolumn{2}{c}{\analogymethod{}} & LRCos \\
			& Inter & IntraL & IntraR & DensRay & SVM & \\
			\hline
			\hline
			capital-common-countries & 0.76& 0.53& 0.56& \textbf{1.00}& \textbf{1.00}& \textbf{1.00}\\
			capital-world & 0.75& 0.44& 0.51& 0.97& 0.96& \textbf{1.00}\\
			city-in-state & 0.71& 0.51& 0.63& 0.78& 0.79& \textbf{0.85}\\
			currency & 0.33& 0.59& 0.48& 0.62& \textbf{0.69}& 0.08\\
			family & 0.84& 0.57& 0.59& 0.91& 0.91& \textbf{0.95}\\
			gram1-adjective-to-adverb & 0.66& 0.50& 0.56& 0.78& 0.84& \textbf{0.88}\\
			gram2-opposite & 0.72& 0.51& 0.54& 0.69& \textbf{0.83}& 0.76\\
			gram3-comparative & 0.75& 0.53& 0.57& 0.84& 0.89& \textbf{0.95}\\
			gram4-superlative & 0.70& 0.53& 0.61& 0.94& \textbf{1.00}& \textbf{1.00}\\
			gram5-present-participle & 0.76& 0.43& 0.48& \textbf{1.00}& 0.94& \textbf{1.00}\\
			gram6-nationality-adjective & 0.70& 0.55& 0.54& 0.90& 0.90& \textbf{0.93}\\
			gram7-past-tense & 0.73& 0.49& 0.47& 0.82& 0.90& \textbf{0.97}\\
			gram8-plural & 0.80& 0.41& 0.42& \textbf{1.00}& \textbf{1.00}& \textbf{1.00}\\
			gram9-plural-verbs & 0.74& 0.43& 0.48& 0.90& 0.97& \textbf{1.00}\\
			\hline
			Micro Mean & 0.73& 0.48& 0.53& 0.88& 0.91& \textbf{0.93}\\
			Macro Mean & 0.71& 0.50& 0.53& 0.87& \textbf{0.90}& 0.89\\
			Macro Std & 0.11& 0.05& 0.06& 0.11& 0.08& \textbf{0.23}\\
		\end{tabular} & 
		\begin{tabular}{l@{\hspace{\sep}}||r@{\hspace{\sep}}r@{\hspace{\sep}}r@{\hspace{\sep}}||r@{\hspace{\sep}}r@{\hspace{\sep}}|r@{\hspace{\sep}}}
			& \multicolumn{3}{c}{Mean Cosine Sim} & \multicolumn{3}{c}{Precision} \\
			& & &  & \multicolumn{2}{c}{\analogymethod{}} & LRCos \\
			& Inter & IntraL & IntraR & DensRay & SVM & \\
			\hline
			\hline
			capital-common-countries & 0.64& 0.37& 0.38& 0.91& \textbf{0.96}& 0.91\\
			capital-world & 0.64& 0.34& 0.36& 0.86& 0.88& \textbf{0.90}\\
			city-in-state & 0.59& 0.37& 0.49& 0.82& 0.85& \textbf{0.87}\\
			currency & 0.37& 0.42& 0.44& \textbf{0.78}& 0.72& 0.56\\
			family & 0.74& 0.48& 0.53& 0.83& \textbf{0.87}& \textbf{0.87}\\
			gram1-adjective-to-adverb & 0.49& 0.21& 0.26& 0.69& \textbf{0.75}& \textbf{0.75}\\
			gram2-opposite & 0.50& 0.24& 0.31& 0.68& \textbf{0.75}& 0.71\\
			gram3-comparative & 0.60& 0.25& 0.40& 0.92& 0.92& \textbf{0.97}\\
			gram4-superlative & 0.54& 0.26& 0.39& \textbf{0.97}& 0.91& 0.94\\
			gram5-present-participle & 0.70& 0.20& 0.21& 0.88& 0.88& \textbf{0.94}\\
			gram6-nationality-adjective & 0.72& 0.41& 0.41& \textbf{0.95}& \textbf{0.95}& 0.93\\
			gram7-past-tense & 0.66& 0.21& 0.22& 0.82& 0.82& \textbf{0.85}\\
			gram8-plural & 0.73& 0.21& 0.23& 0.92& 0.89& \textbf{0.97}\\
			gram9-plural-verbs & 0.64& 0.22& 0.24& 0.87& 0.90& \textbf{0.93}\\
			\hline
			Micro Mean & 0.62& 0.31& 0.36& 0.85& 0.87& \textbf{0.88}\\
			Macro Mean & 0.61& 0.30& 0.35& 0.85& 0.86& \textbf{0.87}\\
			Macro Std & 0.10& 0.09& 0.10& 0.08& 0.07& \textbf{0.11}\\
		\end{tabular}\\[80pt]
		\rotatebox{90}{\textbf{BATS}} &	\begin{tabular}{l@{\hspace{\sep}}||r@{\hspace{\sep}}r@{\hspace{\sep}}r@{\hspace{\sep}}||r@{\hspace{\sep}}r@{\hspace{\sep}}|r@{\hspace{\sep}}}
			& \multicolumn{3}{c}{Mean Cosine Sim} & \multicolumn{3}{c}{Precision} \\
			& & &  & \multicolumn{2}{c}{\analogymethod{}} & LRCos \\
			& Inter & IntraL & IntraR & DensRay & SVM & \\
			\hline
			\hline
			Derivational & 0.63& 0.47& 0.45& 0.74& 0.78& \textbf{0.80}\\
			D01 [noun+less-reg] & 0.51& 0.36& 0.45& 0.50& 0.53& \textbf{0.60}\\
			D02 [un+adj-reg] & 0.71& 0.46& 0.46& 0.68& 0.72& \textbf{0.84}\\
			D03 [adj+ly-reg] & 0.63& 0.50& 0.52& 0.86& 0.88& \textbf{0.90}\\
			D04 [over+adj-reg] & 0.63& 0.45& 0.45& 0.59& \textbf{0.69}& 0.62\\
			D05 [adj+ness-reg] & 0.63& 0.49& 0.51& 0.92& \textbf{1.00}& 0.92\\
			D06 [re+verb-reg] & 0.74& 0.52& 0.49& 0.50& 0.62& \textbf{0.76}\\
			D07 [verb+able-reg] & 0.57& 0.48& 0.45& \textbf{0.71}& 0.66& 0.63\\
			D08 [verb+er-irreg] & 0.55& 0.48& 0.41& 0.84& \textbf{0.88}& 0.79\\
			D09 [verb+tion-irreg] & 0.63& 0.46& 0.41& 0.86& \textbf{0.88}& 0.86\\
			D10 [verb+ment-irreg] & 0.64& 0.46& 0.42& 0.86& 0.88& \textbf{0.90}\\
			\hline
			Encyclopedia & 0.48& 0.43& 0.55& 0.30& 0.43& \textbf{0.45}\\
			E01 [country - capital] & 0.63& 0.47& 0.41& 0.72& 0.96& \textbf{0.98}\\
			E02 [country - language] & 0.40& 0.43& 0.59& 0.24& \textbf{0.35}& 0.33\\
			E03 [UK-city - county] & 0.59& 0.49& 0.57& 0.22& \textbf{0.36}& \textbf{0.36}\\
			E04 [name - nationality] & 0.28& 0.39& 0.64& 0.45& \textbf{0.66}& 0.60\\
			E05 [name - occupation] & 0.44& 0.41& 0.57& 0.42& 0.65& \textbf{0.73}\\
			E06 [animal - young] & 0.47& 0.43& 0.44& 0.05& 0.07& \textbf{0.15}\\
			E07 [animal - sound] & 0.37& 0.43& 0.40& 0.15& 0.20& \textbf{0.22}\\
			E08 [animal - shelter] & 0.44& 0.42& 0.51& 0.00& 0.07& \textbf{0.13}\\
			E09 [things - color] & 0.44& 0.38& 0.81& 0.04& \textbf{0.22}& 0.16\\
			E10 [male - female] & 0.73& 0.43& 0.43& 0.68& 0.68& \textbf{0.78}\\
			\hline
			Inflectional & 0.75& 0.48& 0.51& 0.92& 0.93& \textbf{0.97}\\
			I01 [noun - plural-reg] & 0.79& 0.39& 0.41& 0.98& \textbf{1.00}& \textbf{1.00}\\
			I02 [noun - plural-irreg] & 0.77& 0.40& 0.42& 0.80& 0.80& \textbf{0.84}\\
			I03 [adj - comparative] & 0.75& 0.50& 0.52& 0.97& \textbf{1.00}& \textbf{1.00}\\
			I04 [adj - superlative] & 0.71& 0.51& 0.58& 0.96& 0.96& \textbf{1.00}\\
			I05 [verb-inf - 3pSg] & 0.77& 0.52& 0.53& 0.96& 0.98& \textbf{1.00}\\
			I06 [verb-inf - Ving] & 0.77& 0.51& 0.51& 0.80& 0.88& \textbf{0.96}\\
			I07 [verb-inf - Ved] & 0.75& 0.51& 0.55& 0.92& 0.94& \textbf{1.00}\\
			I08 [verb-Ving - 3pSg] & 0.70& 0.48& 0.51& 0.94& 0.96& \textbf{0.98}\\
			I09 [verb-Ving - Ved] & 0.73& 0.50& 0.54& 0.92& 0.88& \textbf{0.98}\\
			I10 [verb-3pSg - Ved] & 0.72& 0.53& 0.55& 0.96& 0.96& \textbf{1.00}\\
			\hline
			Lexicography & 0.62& 0.37& 0.38& 0.17& 0.20& \textbf{0.26}\\
			L01 [hypernyms - animals] & 0.58& 0.43& 0.53& 0.02& \textbf{0.29}& 0.24\\
			L02 [hypernyms - misc] & 0.55& 0.35& 0.39& \textbf{0.14}& 0.11& \textbf{0.14}\\
			L03 [hyponyms - misc] & 0.63& 0.35& 0.31& \textbf{0.28}& \textbf{0.28}& \textbf{0.28}\\
			L04 [meronyms - substance] & 0.53& 0.36& 0.44& 0.15& \textbf{0.21}& 0.17\\
			L05 [meronyms - member] & 0.58& 0.36& 0.36& 0.10& 0.10& \textbf{0.12}\\
			L06 [meronyms - part] & 0.53& 0.31& 0.30& 0.04& \textbf{0.09}& \textbf{0.09}\\
			L07 [synonyms - intensity] & 0.67& 0.37& 0.37& 0.25& 0.25& \textbf{0.36}\\
			L08 [synonyms - exact] & 0.71& 0.33& 0.31& 0.18& 0.16& \textbf{0.22}\\
			L09 [antonyms - gradable] & 0.68& 0.45& 0.43& 0.35& 0.33& \textbf{0.55}\\
			L10 [antonyms - binary] & 0.72& 0.40& 0.40& 0.18& 0.18& \textbf{0.39}\\
			\hline
			Micro Mean & 0.62& 0.44& 0.47& 0.52& 0.58& \textbf{0.61}\\
			Macro Mean & 0.62& 0.44& 0.47& 0.53& 0.58& \textbf{0.61}\\
			Macro Std & 0.12& 0.06& 0.09& \textbf{0.34}& 0.33& 0.32\\
		\end{tabular}
		&
		\begin{tabular}{l@{\hspace{\sep}}||r@{\hspace{\sep}}r@{\hspace{\sep}}r@{\hspace{\sep}}||r@{\hspace{\sep}}r@{\hspace{\sep}}|r@{\hspace{\sep}}}
			& \multicolumn{3}{c}{Mean Cosine Sim} & \multicolumn{3}{c}{Precision} \\
			& & &  & \multicolumn{2}{c}{\analogymethod{}} & LRCos \\
			& Inter & IntraL & IntraR & DensRay & SVM & \\
			\hline
			\hline
			Derivational & 0.44& 0.21& 0.20& \textbf{0.55}& 0.50& 0.44\\
			D01 [noun+less-reg] & 0.26& 0.16& 0.24& \textbf{0.14}& \textbf{0.14}& 0.05\\
			D02 [un+adj-reg] & 0.47& 0.17& 0.20& \textbf{0.66}& 0.54& 0.58\\
			D03 [adj+ly-reg] & 0.48& 0.17& 0.22& 0.70& \textbf{0.76}& \textbf{0.76}\\
			D04 [over+adj-reg] & 0.39& 0.17& 0.21& 0.41& \textbf{0.44}& 0.30\\
			D05 [adj+ness-reg] & 0.47& 0.21& 0.26& \textbf{0.75}& \textbf{0.75}& 0.65\\
			D06 [re+verb-reg] & 0.56& 0.29& 0.28& \textbf{0.64}& 0.53& 0.58\\
			D07 [verb+able-reg] & 0.38& 0.25& 0.20& \textbf{0.38}& 0.28& 0.19\\
			D08 [verb+er-irreg] & 0.30& 0.24& 0.17& \textbf{0.29}& 0.19& 0.07\\
			D09 [verb+tion-irreg] & 0.51& 0.22& 0.15& \textbf{0.73}& 0.63& 0.51\\
			D10 [verb+ment-irreg] & 0.47& 0.24& 0.15& \textbf{0.60}& 0.56& 0.44\\
			\hline
			Encyclopedia & 0.35& 0.29& 0.42& 0.33& \textbf{0.35}& 0.34\\
			E01 [country - capital] & 0.61& 0.35& 0.32& 0.88& \textbf{0.90}& \textbf{0.90}\\
			E02 [country - language] & 0.36& 0.31& 0.45& \textbf{0.47}& 0.30& 0.36\\
			E03 [UK-city - county] & 0.41& 0.36& 0.52& \textbf{0.14}& \textbf{0.14}& \textbf{0.14}\\
			E04 [name - nationality] & 0.20& 0.20& 0.39& \textbf{0.33}& \textbf{0.33}& 0.26\\
			E05 [name - occupation] & 0.33& 0.21& 0.40& 0.45& \textbf{0.62}& 0.52\\
			E06 [animal - young] & 0.34& 0.36& 0.38& 0.06& 0.06& \textbf{0.12}\\
			E07 [animal - sound] & 0.15& 0.31& 0.25& \textbf{0.17}& 0.03& 0.00\\
			E08 [animal - shelter] & 0.25& 0.29& 0.39& 0.00& \textbf{0.16}& 0.09\\
			E09 [things - color] & 0.20& 0.23& 0.63& 0.08& 0.15& \textbf{0.21}\\
			E10 [male - female] & 0.62& 0.28& 0.33& 0.66& \textbf{0.68}& \textbf{0.68}\\
			\hline
			Inflectional & 0.63& 0.22& 0.23& \textbf{0.88}& 0.87& \textbf{0.88}\\
			I01 [noun - plural-reg] & 0.69& 0.13& 0.16& 0.84& 0.84& \textbf{0.88}\\
			I02 [noun - plural-irreg] & 0.62& 0.12& 0.16& 0.67& 0.69& \textbf{0.75}\\
			I03 [adj - comparative] & 0.63& 0.23& 0.37& 0.97& 0.97& \textbf{1.00}\\
			I04 [adj - superlative] & 0.59& 0.26& 0.39& 0.93& 0.93& \textbf{0.97}\\
			I05 [verb-inf - 3pSg] & 0.65& 0.26& 0.33& \textbf{1.00}& \textbf{1.00}& \textbf{1.00}\\
			I06 [verb-inf - Ving] & 0.67& 0.26& 0.19& \textbf{0.84}& 0.82& 0.82\\
			I07 [verb-inf - Ved] & 0.66& 0.25& 0.20& \textbf{0.92}& 0.90& 0.88\\
			I08 [verb-Ving - 3pSg] & 0.56& 0.17& 0.31& 0.90& \textbf{0.92}& 0.90\\
			I09 [verb-Ving - Ved] & 0.64& 0.18& 0.19& \textbf{0.84}& \textbf{0.84}& 0.82\\
			I10 [verb-3pSg - Ved] & 0.62& 0.33& 0.20& \textbf{0.90}& 0.88& 0.88\\
			\hline
			Lexicography & 0.45& 0.17& 0.18& \textbf{0.19}& 0.17& 0.18\\
			L01 [hypernyms - animals] & 0.47& 0.32& 0.47& 0.00& \textbf{0.05}& \textbf{0.05}\\
			L02 [hypernyms - misc] & 0.42& 0.21& 0.21& \textbf{0.29}& 0.21& 0.10\\
			L03 [hyponyms - misc] & 0.52& 0.15& 0.15& \textbf{0.19}& 0.14& 0.14\\
			L04 [meronyms - substance] & 0.35& 0.16& 0.24& \textbf{0.15}& 0.09& 0.11\\
			L05 [meronyms - member] & 0.36& 0.15& 0.15& \textbf{0.10}& 0.08& 0.08\\
			L06 [meronyms - part] & 0.34& 0.14& 0.12& \textbf{0.09}& \textbf{0.09}& 0.02\\
			L07 [synonyms - intensity] & 0.51& 0.17& 0.16& 0.24& 0.26& \textbf{0.30}\\
			L08 [synonyms - exact] & 0.55& 0.11& 0.11& 0.15& 0.15& \textbf{0.22}\\
			L09 [antonyms - gradable] & 0.45& 0.18& 0.19& 0.41& 0.41& \textbf{0.43}\\
			L10 [antonyms - binary] & 0.50& 0.14& 0.15& 0.21& 0.17& \textbf{0.31}\\
			\hline
			Micro Mean & 0.47& 0.22& 0.26& \textbf{0.49}& 0.48& 0.47\\
			Macro Mean & 0.46& 0.22& 0.26& \textbf{0.48}& 0.47& 0.45\\
			Macro Std & 0.14& 0.07& 0.12& 0.31& 0.31& \textbf{0.32}\\
		\end{tabular}
	\end{tabular}
	\caption{\centering Detailed results for all combinations of FastText/Google News embeddings and Google Analogy and BATS analogies. In this table the cosine similarity is computed in the orthogonal complement. See the main paper for more details.\tablabel{complement}}
\end{table*}

\begin{table*}
	\def\sep{0.01cm}
	\tiny
	\centering
	\begin{tabular}{ccc}
		& \textbf{FastText} &\textbf{ Google News}\\[10pt]
		
		\rotatebox{90}{\textbf{Google Analogy}} &	\begin{tabular}{l@{\hspace{\sep}}||r@{\hspace{\sep}}r@{\hspace{\sep}}r@{\hspace{\sep}}||r@{\hspace{\sep}}r@{\hspace{\sep}}|r@{\hspace{\sep}}}
			& \multicolumn{3}{c}{Mean Cosine Sim} & \multicolumn{3}{c}{Precision} \\
			& & &  & \multicolumn{2}{c}{\analogymethod{}} & LRCos \\
			& Inter & IntraL & IntraR & DensRay & SVM & \\
			\hline
			\hline
			capital-common-countries & 0.76& 0.53& 0.56& \textbf{1.00}& \textbf{1.00}& \textbf{1.00}\\
			capital-world & 0.75& 0.44& 0.51& 0.98& 0.98& \textbf{1.00}\\
			city-in-state & 0.71& 0.51& 0.63& \textbf{0.87}& 0.85& 0.85\\
			currency & 0.33& 0.59& 0.48& \textbf{0.54}& 0.31& 0.08\\
			family & 0.84& 0.57& 0.59& 0.91& 0.91& \textbf{0.95}\\
			gram1-adjective-to-adverb & 0.66& 0.50& 0.56& \textbf{0.88}& 0.84& \textbf{0.88}\\
			gram2-opposite & 0.72& 0.51& 0.54& 0.72& \textbf{0.83}& 0.76\\
			gram3-comparative & 0.75& 0.53& 0.57& 0.92& \textbf{0.95}& \textbf{0.95}\\
			gram4-superlative & 0.70& 0.53& 0.61& \textbf{1.00}& \textbf{1.00}& \textbf{1.00}\\
			gram5-present-participle & 0.76& 0.43& 0.48& \textbf{1.00}& \textbf{1.00}& \textbf{1.00}\\
			gram6-nationality-adjective & 0.70& 0.55& 0.54& \textbf{0.93}& \textbf{0.93}& \textbf{0.93}\\
			gram7-past-tense & 0.73& 0.49& 0.47& 0.95& 0.93& \textbf{0.97}\\
			gram8-plural & 0.80& 0.41& 0.42& \textbf{1.00}& \textbf{1.00}& \textbf{1.00}\\
			gram9-plural-verbs & 0.74& 0.43& 0.48& \textbf{1.00}& \textbf{1.00}& \textbf{1.00}\\
			\hline
			Micro Mean & 0.73& 0.48& 0.53& \textbf{0.93}& 0.92& \textbf{0.93}\\
			Macro Mean & 0.71& 0.50& 0.53& \textbf{0.91}& 0.90& 0.89\\
			Macro Std & 0.11& 0.05& 0.06& 0.12& 0.17& \textbf{0.23}\\
		\end{tabular} & 
		\begin{tabular}{l@{\hspace{\sep}}||r@{\hspace{\sep}}r@{\hspace{\sep}}r@{\hspace{\sep}}||r@{\hspace{\sep}}r@{\hspace{\sep}}|r@{\hspace{\sep}}}
			& \multicolumn{3}{c}{Mean Cosine Sim} & \multicolumn{3}{c}{Precision} \\
			& & &  & \multicolumn{2}{c}{\analogymethod{}} & LRCos \\
			& Inter & IntraL & IntraR & DensRay & SVM & \\
			\hline
			\hline
			capital-common-countries & 0.64& 0.37& 0.38& \textbf{0.91}& \textbf{0.91}& \textbf{0.91}\\
			capital-world & 0.64& 0.34& 0.36& \textbf{0.90}& \textbf{0.90}& \textbf{0.90}\\
			city-in-state & 0.59& 0.37& 0.49& 0.85& \textbf{0.88}& 0.87\\
			currency & 0.37& 0.42& 0.44& \textbf{0.67}& 0.50& 0.56\\
			family & 0.74& 0.48& 0.53& \textbf{0.87}& \textbf{0.87}& \textbf{0.87}\\
			gram1-adjective-to-adverb & 0.49& 0.21& 0.26& \textbf{0.78}& 0.75& 0.75\\
			gram2-opposite & 0.50& 0.24& 0.31& 0.71& \textbf{0.75}& 0.71\\
			gram3-comparative & 0.60& 0.25& 0.40& \textbf{0.97}& 0.95& \textbf{0.97}\\
			gram4-superlative & 0.54& 0.26& 0.39& \textbf{0.94}& \textbf{0.94}& \textbf{0.94}\\
			gram5-present-participle & 0.70& 0.20& 0.21& \textbf{0.94}& 0.91& \textbf{0.94}\\
			gram6-nationality-adjective & 0.72& 0.41& 0.41& \textbf{0.93}& \textbf{0.93}& \textbf{0.93}\\
			gram7-past-tense & 0.66& 0.21& 0.22& \textbf{0.88}& 0.85& 0.85\\
			gram8-plural & 0.73& 0.21& 0.23& \textbf{0.97}& 0.89& \textbf{0.97}\\
			gram9-plural-verbs & 0.64& 0.22& 0.24& \textbf{0.93}& 0.87& \textbf{0.93}\\
			\hline
			Micro Mean & 0.62& 0.31& 0.36& \textbf{0.89}& 0.87& 0.88\\
			Macro Mean & 0.61& 0.30& 0.35& \textbf{0.88}& 0.85& 0.87\\
			Macro Std & 0.10& 0.09& 0.10& 0.09& \textbf{0.11}& 0.11\\
		\end{tabular}\\[80pt]
		\rotatebox{90}{\textbf{BATS}} &	\begin{tabular}{l@{\hspace{\sep}}||r@{\hspace{\sep}}r@{\hspace{\sep}}r@{\hspace{\sep}}||r@{\hspace{\sep}}r@{\hspace{\sep}}|r@{\hspace{\sep}}}
			& \multicolumn{3}{c}{Mean Cosine Sim} & \multicolumn{3}{c}{Precision} \\
			& & &  & \multicolumn{2}{c}{\analogymethod{}} & LRCos \\
			& Inter & IntraL & IntraR & DensRay & SVM & \\
			\hline
			\hline
			Derivational & 0.63& 0.47& 0.45& \textbf{0.81}& 0.80& 0.80\\
			D01 [noun+less-reg] & 0.51& 0.36& 0.45& \textbf{0.60}& \textbf{0.60}& \textbf{0.60}\\
			D02 [un+adj-reg] & 0.71& 0.46& 0.46& 0.80& 0.72& \textbf{0.84}\\
			D03 [adj+ly-reg] & 0.63& 0.50& 0.52& \textbf{0.90}& \textbf{0.90}& \textbf{0.90}\\
			D04 [over+adj-reg] & 0.63& 0.45& 0.45& 0.66& \textbf{0.69}& 0.62\\
			D05 [adj+ness-reg] & 0.63& 0.49& 0.51& \textbf{1.00}& 0.96& 0.92\\
			D06 [re+verb-reg] & 0.74& 0.52& 0.49& 0.59& 0.71& \textbf{0.76}\\
			D07 [verb+able-reg] & 0.57& 0.48& 0.45& \textbf{0.74}& 0.63& 0.63\\
			D08 [verb+er-irreg] & 0.55& 0.48& 0.41& \textbf{0.91}& 0.86& 0.79\\
			D09 [verb+tion-irreg] & 0.63& 0.46& 0.41& 0.91& \textbf{0.93}& 0.86\\
			D10 [verb+ment-irreg] & 0.64& 0.46& 0.42& \textbf{0.92}& 0.90& 0.90\\
			\hline
			Encyclopedia & 0.48& 0.43& 0.55& 0.41& 0.43& \textbf{0.45}\\
			E01 [country - capital] & 0.63& 0.47& 0.41& 0.96& 0.96& \textbf{0.98}\\
			E02 [country - language] & 0.40& 0.43& 0.59& \textbf{0.33}& 0.24& \textbf{0.33}\\
			E03 [UK-city - county] & 0.59& 0.49& 0.57& 0.30& \textbf{0.42}& 0.36\\
			E04 [name - nationality] & 0.28& 0.39& 0.64& 0.49& 0.51& \textbf{0.60}\\
			E05 [name - occupation] & 0.44& 0.41& 0.57& \textbf{0.75}& 0.73& 0.73\\
			E06 [animal - young] & 0.47& 0.43& 0.44& 0.10& \textbf{0.15}& \textbf{0.15}\\
			E07 [animal - sound] & 0.37& 0.43& 0.40& \textbf{0.22}& 0.17& \textbf{0.22}\\
			E08 [animal - shelter] & 0.44& 0.42& 0.51& 0.02& \textbf{0.13}& \textbf{0.13}\\
			E09 [things - color] & 0.44& 0.38& 0.81& 0.12& \textbf{0.22}& 0.16\\
			E10 [male - female] & 0.73& 0.43& 0.43& 0.76& 0.71& \textbf{0.78}\\
			\hline
			Inflectional & 0.75& 0.48& 0.51& \textbf{0.97}& \textbf{0.97}& \textbf{0.97}\\
			I01 [noun - plural-reg] & 0.79& 0.39& 0.41& \textbf{1.00}& \textbf{1.00}& \textbf{1.00}\\
			I02 [noun - plural-irreg] & 0.77& 0.40& 0.42& \textbf{0.84}& 0.82& \textbf{0.84}\\
			I03 [adj - comparative] & 0.75& 0.50& 0.52& \textbf{1.00}& \textbf{1.00}& \textbf{1.00}\\
			I04 [adj - superlative] & 0.71& 0.51& 0.58& 0.96& 0.96& \textbf{1.00}\\
			I05 [verb-inf - 3pSg] & 0.77& 0.52& 0.53& \textbf{1.00}& \textbf{1.00}& \textbf{1.00}\\
			I06 [verb-inf - Ving] & 0.77& 0.51& 0.51& \textbf{0.96}& \textbf{0.96}& \textbf{0.96}\\
			I07 [verb-inf - Ved] & 0.75& 0.51& 0.55& \textbf{1.00}& \textbf{1.00}& \textbf{1.00}\\
			I08 [verb-Ving - 3pSg] & 0.70& 0.48& 0.51& \textbf{0.98}& \textbf{0.98}& \textbf{0.98}\\
			I09 [verb-Ving - Ved] & 0.73& 0.50& 0.54& \textbf{0.98}& 0.96& \textbf{0.98}\\
			I10 [verb-3pSg - Ved] & 0.72& 0.53& 0.55& \textbf{1.00}& \textbf{1.00}& \textbf{1.00}\\
			\hline
			Lexicography & 0.62& 0.37& 0.38& 0.21& 0.22& \textbf{0.26}\\
			L01 [hypernyms - animals] & 0.58& 0.43& 0.53& 0.20& \textbf{0.37}& 0.24\\
			L02 [hypernyms - misc] & 0.55& 0.35& 0.39& \textbf{0.23}& 0.16& 0.14\\
			L03 [hyponyms - misc] & 0.63& 0.35& 0.31& \textbf{0.33}& 0.28& 0.28\\
			L04 [meronyms - substance] & 0.53& 0.36& 0.44& 0.10& 0.15& \textbf{0.17}\\
			L05 [meronyms - member] & 0.58& 0.36& 0.36& \textbf{0.12}& 0.10& \textbf{0.12}\\
			L06 [meronyms - part] & 0.53& 0.31& 0.30& 0.04& \textbf{0.15}& 0.09\\
			L07 [synonyms - intensity] & 0.67& 0.37& 0.37& 0.27& 0.32& \textbf{0.36}\\
			L08 [synonyms - exact] & 0.71& 0.33& 0.31& 0.18& 0.16& \textbf{0.22}\\
			L09 [antonyms - gradable] & 0.68& 0.45& 0.43& 0.43& 0.37& \textbf{0.55}\\
			L10 [antonyms - binary] & 0.72& 0.40& 0.40& 0.18& 0.18& \textbf{0.39}\\
			\hline
			Micro Mean & 0.62& 0.44& 0.47& 0.59& 0.60& \textbf{0.61}\\
			Macro Mean & 0.62& 0.44& 0.47& 0.60& 0.60& \textbf{0.61}\\
			Macro Std & 0.12& 0.06& 0.09& \textbf{0.34}& 0.33& 0.32\\
		\end{tabular}
		&
		\begin{tabular}{l@{\hspace{\sep}}||r@{\hspace{\sep}}r@{\hspace{\sep}}r@{\hspace{\sep}}||r@{\hspace{\sep}}r@{\hspace{\sep}}|r@{\hspace{\sep}}}
			& \multicolumn{3}{c}{Mean Cosine Sim} & \multicolumn{3}{c}{Precision} \\
			& & &  & \multicolumn{2}{c}{\analogymethod{}} & LRCos \\
			& Inter & IntraL & IntraR & DensRay & SVM & \\
			\hline
			\hline
			Derivational & 0.44& 0.21& 0.20& \textbf{0.51}& 0.48& 0.44\\
			D01 [noun+less-reg] & 0.26& 0.16& 0.24& \textbf{0.10}& \textbf{0.10}& 0.05\\
			D02 [un+adj-reg] & 0.47& 0.17& 0.20& \textbf{0.66}& 0.58& 0.58\\
			D03 [adj+ly-reg] & 0.48& 0.17& 0.22& 0.76& \textbf{0.78}& 0.76\\
			D04 [over+adj-reg] & 0.39& 0.17& 0.21& \textbf{0.41}& \textbf{0.41}& 0.30\\
			D05 [adj+ness-reg] & 0.47& 0.21& 0.26& \textbf{0.75}& 0.70& 0.65\\
			D06 [re+verb-reg] & 0.56& 0.29& 0.28& \textbf{0.69}& 0.61& 0.58\\
			D07 [verb+able-reg] & 0.38& 0.25& 0.20& \textbf{0.28}& 0.22& 0.19\\
			D08 [verb+er-irreg] & 0.30& 0.24& 0.17& \textbf{0.12}& 0.10& 0.07\\
			D09 [verb+tion-irreg] & 0.51& 0.22& 0.15& \textbf{0.61}& \textbf{0.61}& 0.51\\
			D10 [verb+ment-irreg] & 0.47& 0.24& 0.15& \textbf{0.54}& 0.50& 0.44\\
			\hline
			Encyclopedia & 0.35& 0.29& 0.42& \textbf{0.35}& 0.32& 0.34\\
			E01 [country - capital] & 0.61& 0.35& 0.32& \textbf{0.90}& \textbf{0.90}& \textbf{0.90}\\
			E02 [country - language] & 0.36& 0.31& 0.45& \textbf{0.43}& 0.21& 0.36\\
			E03 [UK-city - county] & 0.41& 0.36& 0.52& \textbf{0.14}& 0.12& \textbf{0.14}\\
			E04 [name - nationality] & 0.20& 0.20& 0.39& \textbf{0.26}& 0.15& \textbf{0.26}\\
			E05 [name - occupation] & 0.33& 0.21& 0.40& \textbf{0.60}& 0.57& 0.52\\
			E06 [animal - young] & 0.34& 0.36& 0.38& 0.09& \textbf{0.12}& \textbf{0.12}\\
			E07 [animal - sound] & 0.15& 0.31& 0.25& \textbf{0.08}& 0.00& 0.00\\
			E08 [animal - shelter] & 0.25& 0.29& 0.39& 0.07& \textbf{0.14}& 0.09\\
			E09 [things - color] & 0.20& 0.23& 0.63& 0.19& 0.19& \textbf{0.21}\\
			E10 [male - female] & 0.62& 0.28& 0.33& \textbf{0.68}& \textbf{0.68}& \textbf{0.68}\\
			\hline
			Inflectional & 0.63& 0.22& 0.23& \textbf{0.88}& \textbf{0.88}& \textbf{0.88}\\
			I01 [noun - plural-reg] & 0.69& 0.13& 0.16& 0.86& 0.84& \textbf{0.88}\\
			I02 [noun - plural-irreg] & 0.62& 0.12& 0.16& 0.69& 0.71& \textbf{0.75}\\
			I03 [adj - comparative] & 0.63& 0.23& 0.37& 0.97& \textbf{1.00}& \textbf{1.00}\\
			I04 [adj - superlative] & 0.59& 0.26& 0.39& \textbf{0.97}& \textbf{0.97}& \textbf{0.97}\\
			I05 [verb-inf - 3pSg] & 0.65& 0.26& 0.33& \textbf{1.00}& \textbf{1.00}& \textbf{1.00}\\
			I06 [verb-inf - Ving] & 0.67& 0.26& 0.19& \textbf{0.86}& 0.82& 0.82\\
			I07 [verb-inf - Ved] & 0.66& 0.25& 0.20& \textbf{0.92}& 0.90& 0.88\\
			I08 [verb-Ving - 3pSg] & 0.56& 0.17& 0.31& \textbf{0.90}& \textbf{0.90}& \textbf{0.90}\\
			I09 [verb-Ving - Ved] & 0.64& 0.18& 0.19& \textbf{0.84}& 0.82& 0.82\\
			I10 [verb-3pSg - Ved] & 0.62& 0.33& 0.20& \textbf{0.88}& \textbf{0.88}& \textbf{0.88}\\
			\hline
			Lexicography & 0.45& 0.17& 0.18& \textbf{0.19}& 0.17& 0.18\\
			L01 [hypernyms - animals] & 0.47& 0.32& 0.47& \textbf{0.08}& \textbf{0.08}& 0.05\\
			L02 [hypernyms - misc] & 0.42& 0.21& 0.21& \textbf{0.26}& \textbf{0.26}& 0.10\\
			L03 [hyponyms - misc] & 0.52& 0.15& 0.15& \textbf{0.19}& 0.14& 0.14\\
			L04 [meronyms - substance] & 0.35& 0.16& 0.24& 0.09& 0.09& \textbf{0.11}\\
			L05 [meronyms - member] & 0.36& 0.15& 0.15& \textbf{0.10}& 0.06& 0.08\\
			L06 [meronyms - part] & 0.34& 0.14& 0.12& \textbf{0.09}& \textbf{0.09}& 0.02\\
			L07 [synonyms - intensity] & 0.51& 0.17& 0.16& 0.26& 0.26& \textbf{0.30}\\
			L08 [synonyms - exact] & 0.55& 0.11& 0.11& 0.15& 0.15& \textbf{0.22}\\
			L09 [antonyms - gradable] & 0.45& 0.18& 0.19& \textbf{0.45}& 0.41& 0.43\\
			L10 [antonyms - binary] & 0.50& 0.14& 0.15& 0.19& 0.17& \textbf{0.31}\\
			\hline
			Micro Mean & 0.47& 0.22& 0.26& \textbf{0.49}& 0.47& 0.47\\
			Macro Mean & 0.46& 0.22& 0.26& \textbf{0.48}& 0.46& 0.45\\
			Macro Std & 0.14& 0.07& 0.12& 0.32& \textbf{0.32}& 0.32\\
		\end{tabular}
	\end{tabular}
	\caption{\centering Detailed results for all combinations of FastText/Google News embeddings and Google Analogy and BATS analogies. In this table the cosine similarity is computed in the original space. See the main paper for more details. \tablabel{original}}
\end{table*}

\end{document}